\documentclass[10pt,twocolumn,letterpaper]{article}
\usepackage{cvpr}              

\usepackage{graphicx}
\usepackage{amsmath}
\usepackage{amssymb}
\usepackage{booktabs}
\usepackage{multirow}
\usepackage{graphicx}

\usepackage[pagebackref,breaklinks,colorlinks]{hyperref}

\usepackage[capitalize]{cleveref}
\crefname{section}{Sec.}{Secs.}
\Crefname{section}{Section}{Sections}
\Crefname{table}{Table}{Tables}
\crefname{table}{Tab.}{Tabs.}

\def\cvprPaperID{4520} 
\def\confName{CVPR}
\def\confYear{2022}

\begin{document}

\title{From Representation to Reasoning: Towards both Evidence and Commonsense Reasoning for Video Question-Answering}

\author{\textnormal{Jiangtong Li}$^{1}$, \textnormal{Li Niu}$^{1}$\thanks{Corresponding author.}~, \textnormal{Liqing Zhang}$^{1*}$ \\
$^{1}$ Department of Computer Science and Engineering, MoE Key Lab of Artificial Intelligence, \\
       Shanghai Jiao Tong University\\
{\tt\small \{keep\_moving-Lee,ustcnewly,lqzhang\}@sjtu.edu.cn}
}
\maketitle

\begin{abstract}
Video understanding has achieved great success in representation learning, such as video caption, video object grounding, and video descriptive question-answer.
However, current methods still struggle on video reasoning, including evidence reasoning and commonsense reasoning.
To facilitate deeper video understanding towards video reasoning, we present the task of Causal-VidQA, which includes four types of questions ranging from scene description (description) to evidence reasoning (explanation) and commonsense reasoning (prediction and counterfactual). 
For commonsense reasoning, we set up a two-step solution by answering the question and providing a proper reason.
Through extensive experiments on existing VideoQA methods, we find that the state-of-the-art methods are strong in descriptions but weak in reasoning.
We hope that Causal-VidQA can guide the research of video understanding from representation learning to deeper reasoning.
The dataset and related resources are available at  \url{https://github.com/bcmi/Causal-VidQA.git}.
\end{abstract}

\section{Introduction}
\label{sec:intro}

Videos, which are organized as a sequence of images along the temporal dimension, usually contain richer temporal and causal relations than simple images \cite{buchsbaum2015inferring}. 
With advanced neural network, video understanding has achieved great progress in representation learning, such as video captioning \cite{KrishnaHRFN17}, video action recognition \cite{SenerSY20}, video relation grounding \cite{XiaoSYTC20}, video descriptive question-answering \cite{XuMYR16}, and video instance segmentation \cite{YangFX19}. 
Therefore, for a computational model, recognizing some independent actions or segmenting some specific instances in a video becomes relatively easy \cite{HeZRS16, LinGH19, ZhouZCSX18}, however, performing the reasoning from video clips remains a great challenge. 
On the contrary, it is easy for human beings to answer reasoning questions from video clips, such as explaining why something is happening, predicting what is about to happen, and imagining what would happen under different conditions \cite{spelke2000core}. 

\begin{figure}
\begin{center}
\includegraphics[width=1.0\linewidth]{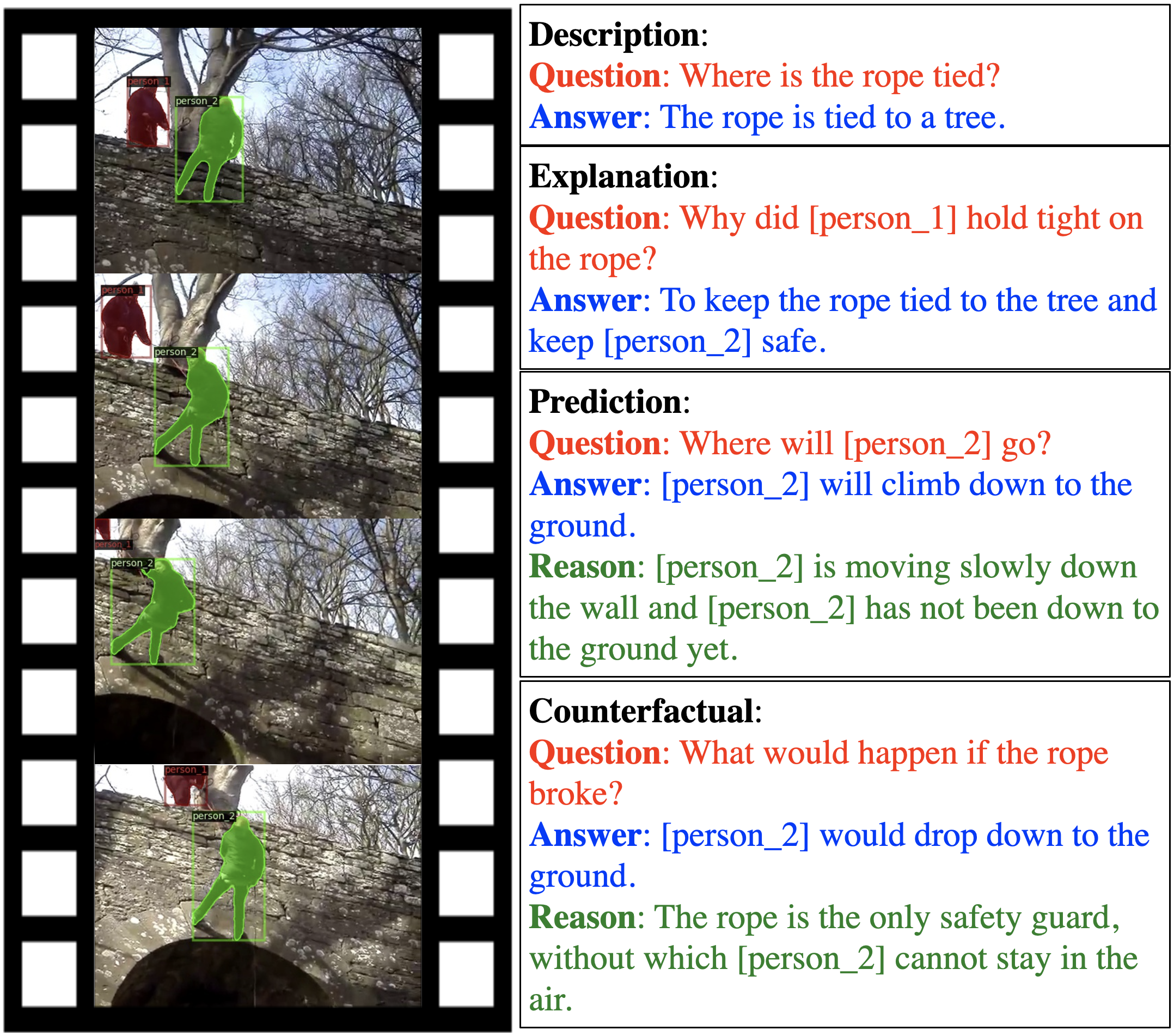}
\end{center}
   \caption{Sample video, questions, and answers from our Causal-VidQA dataset. 
   Causal-VidQA is designed to evaluate whether models can understand what is in the video (description), 
   explain the intentions of actions or procedures to certain targets (explanation), 
   predict what will happen in the future (prediction), and imagine the scenarios in different conditions (counterfactual). }
\label{fig:example}
\vspace{-10pt}
\end{figure}

For example, in the video clip in Figure \ref{fig:example}, a man is \textit{climbing down along the ropes}. 
Recognizing the human actions like ``\textit{abseiling}'' or segmenting and tracking ``\textit{[person\_1]}'' is easy for the state-of-the-art vision systems \cite{Feichtenhofer0M19, YangW020}.
Nevertheless, for the evidence reasoning (explanatory) and commonsense reasoning (prediction and counterfactual) (\eg \textit{Why did [person\_1] hold tight on the rope?} \textit{Where will [person\_2] go?} and \textit{What would happen if the rope broke?}), human beings are capable of correlating \textit{the position and tendency of [person\_1] and [person\_2]} to answer the aforementioned questions with proper reasons, but current models still struggle on the reasoning tasks \cite{XiaoSYC21, YiGLK0TT20}.
Considering that most of the video tasks mainly focus on representation learning \cite{KrishnaHRFN17,SenerSY20,XiaoSYTC20} and reasoning is also less explored, now is the right time to explore video reasoning.

Usually visual reasoning can be divided into two groups, evidence reasoning (\ie all the clues to the answer are visible in visual content) and commonsense reasoning (\ie some clues to the answer need to be imagined beyond the visual content).
Based on the definition, some works have taken the visual reasoning into consideration.
As an image dataset for visual reasoning, VCR \cite{ZellersBFC19} aims to provide a proper reason while answering a certain question, however, images do not include enough temporal relation and action for commonsense reason. 
There are also some video datasets for visual reasoning.
Social-IQ \cite{0001CLTM19} focuses on comprehending complex human social behaviors with rich causal inference, but the scope is quite limited and the scale is also too small (only 1250 videos with 7500 question-answer pairs).
CLEVRER \cite{YiGLK0TT20} focuses on
the causal relations grounded in object dynamics and physical interactions, but it only targets at the virtual scene and ignores the reasoning in reality.
NExT-QA \cite{XiaoSYC21} focuses on the causal and temporal action while ensuring that the answers can be inferred from the video clips. 
However, NExT-QA \cite{XiaoSYC21} only focuses on evidence reasoning, where an appropriate expansion (\ie commonsense reasoning) is missing.

To facilitate a deeper understanding towards video reasoning, we present the task of Causal-VidQA. 
Given a video clip, our Causal-VidQA task requires the model to answer four types of questions including scene description (description), evidence reasoning (explanation), and commonsense reasoning (prediction and counterfactual) to thoroughly understand the video content.
Moreover, for the commonsense reasoning questions (\ie prediction and counterfactual), the model is required to not only provide a right answer but also offer a proper reason justifying why that answer is true, referring to some video details and commonsense knowledge.
Our new dataset Causal-VidQA has 26,900 unique video clips and 107,600 question-answer pairs including descriptive, explanatory, predictive, and counterfactual questions, which makes Causal-VidQA the first large-scale dataset in this realm.
Different from existing datasets, our dataset focuses on evidence and commonsense reasoning in real-world actions, including large-scale action categories and various types of questions to satisfy the requirement of deeper video understanding.
Considering that a reasoning question may correspond to more than one rational answers and reasons, all our tasks are framed as multiple-choice question-answering.

When constructing our dataset, we consider two critical problems.
First, the number of action categories should be large enough to prevent learning a short-cut for reasoning by correlating the action categories with questions and answers.
Therefore, we study several different video datasets and finally decide to use the Kinetics-700 \cite{KayCSZHVVGBNSZ17} as our video dataset, which includes 700 different action categories in real world.
Besides, we also split the training/validation/testing set by video action category.
Second, the instances in video clip should be described precisely and briefly to ensure that the core of our task is Causal-VidQA instead of video object grounding.
To solve this problem, we combine the image instance segmentation and video instance segmentation together to replace the text description with some explicit references to video regions for all frames, like ``\textit{[person\_1]}'' and ``\textit{[person\_1]}'' in Figure \ref{fig:example}. 
Detailed process of dataset construction is in Sec. \ref{sec:dataset:construction}.


Based on Causal-VidQA, we evaluate different state-of-the-art VideoQA methods \cite{FanZZW0H19, GaoGCN18, JangSYKK17, JiangH20, LeLV020, HuangCZDTG20, ParkLS21}. 
Although some methods achieve satisfied results on the descriptive and explanatory questions, their performances on predictive, and counterfactual questions are far from satisfactory. 
These experimental results represents that these models do not truly understand the causal relation and fail to reason about commonsense phenomena. 
Hence, Causal-VidQA offers new challenges for deeper video understanding.
Our contributions can be summarized as 
\begin{itemize}
    \item We explore evidence and commonsense reasoning to advance VideoQA beyond representation learning towards deeper reasoning;
    \item We contribute Causal-VidQA, a new challenging VideoQA benchmark containing four types of questions (\ie description, explanation, prediction, and counterfactual);
    \item We extensively evaluate some SOTA video reasoning methods on our Causal-VidQA dataset, providing detailed comparison and in-depth analyses.
\end{itemize}

\section{Related Work}
\label{sec:related}

Our work can be related to two recent research directions: language-guided video understanding benchmark and visual question-answering technique. 

\subsection{Language-guided Video Understanding}

With rapid growth of video data on the Internet \cite{HeilbronEGN15, KayCSZHVVGBNSZ17}, language-guided video understanding tasks have received considerable interest in recent years.
Early in this field, researchers focused more on video captioning \cite{GuadarramaKMVMDS13}, localizing video segments from natural language queries \cite{HendricksWSSDR17}, and descriptive video question-answering \cite{LiSCTGJL16, MaharajBRCP17, XuMYR16}. 
They mainly required representation learning towards the instances and actions to accomplish the corresponding tasks.

Recently, researchers have explored different approaches to various video reasoning tasks.
Among them, TGIF-QA \cite{JangSYKK17} and ActivityNet-QA \cite{YuXYYZZT19} manually annotated short and long web video respectively to explore the evidence reasoning, especially spatio-temporal reasoning. 
However, the videos in TGIF-QA are usually too short (less than 3s) and the scale of ActivityNet-QA is also relatively small.
MovieQA \cite{TapaswiZSTUF16}, TVQA \cite{LeiYBB18} invoked causal and temporal questions based on movies or TV shows. 
The answers in MovieQA are biased towards textual plot understanding \cite{XiaoSYC21}, and the QA pairs in TVQA are biased towards the actor dialogue comprehension \cite{WinterbottomXMM20}, which severely diminishes their challenge for visual reasoning \cite{JiangH20, LeLV020}.
Different from them, which mainly focused on spatio-temporal relation reasoning in videos, our Causal-VidQA dataset explores the evidence reasoning from general explanation on real-world actions and targets at a richer set of actions categories in daily life.

Some works also explored the commonsense reasoning based on images \cite{VondrickOPT16, ZellersBFC19}.
For example, Motivation \cite{VondrickOPT16} aimed to predict the motivation of a certain action in static images.
VCR \cite{ZellersBFC19} targeted at commonsense reasoning from static images in movie with a two-step solution (answer-reason).
Our dataset is essentially different in that we require the models to speculate or make commonsense reasoning in daily videos towards real-world actions and interactions, where the static images \cite{VondrickOPT16, ZellersBFC19} cannot include enough dynamic information for commonsense reasoning and movie scenes \cite{ZellersBFC19} are also far from daily life.

Another type of works focused on physical world reasoning, which usually collect data in simulated environments.
For example, COG \cite{YangGWSS18} and MarioQA \cite{MunSJH17} used simulated environments to generate synthetic data. 
As an extension, CLEVRER \cite{YiGLK0TT20} focused on the causal relations grounded in physical world and also emphasized compositionality in logic context.
Our work differs in that we explore video reasoning from daily videos towards real-world actions and interactions.
Social-IQ \cite{0001CLTM19} was a newly proposed benchmark and discussed causal relations in human social interactions from three modalities (video, transcript, and audio).
However, this dataset is small-scale and only focuses on limited scenes with three modalities (video, transcript, and audio).
Recently, V2C \cite{FangGBBY20} and NExT-QA \cite{XiaoSYC21} explored video understanding from commonsense caption and evidence reasoning. 
Compared with V2C \cite{FangGBBY20}, in our work, the commonsense reasoning is based on specific question and is explored in a two-step manner (answer-reason).
Besides, our commonsense reasoning is based on real-world videos instead of movie clips, which makes our target closer to daily life.
NExT-QA \cite{XiaoSYC21} focused on the temporal structure of actions and evidence reasoning, which ensures that the answers to the questions can be found in video contents.
As a complementary research direction of NExT-QA \cite{XiaoSYC21}, Causal-VidQA further emphasizes commonsense to imagine the potential answers and reasons. 

\subsection{Video Question-Answering Technique}

Visual question-answering can be divided into three parts: video representation, text representation and information fusion between video and text.
The video representation has been propelled by the advancements in image classification \cite{HeZRS16}, object detection \cite{RenHGS15}, and video action recognition \cite{CarreiraZ17}.
Existing works \cite{FanZZW0H19,LeLV020,LiSGLH0G19,ParkLS21} usually applied 2D convolutional neural network (CNN) (\eg, ResNet \cite{HeZRS16}, and Faster R-CNN \cite{RenHGS15}) to extract frame-level appearance features or object features, 3D CNN (\eg, C3D \cite{TranBFTP15}, I3D \cite{CarreiraZ17, HaraKS18}) to extract clip-level motion features.
The text representation is also driven by the improvement of word embedding \cite{PenningtonSM14} and pre-trained language model \cite{VaswaniSPUJGKP17, DevlinCLT19}.

According to the fashion of information fusion between video and text, existing approaches can be categorized as temporal attention and spatio-temporal attention.
For the temporal attention,  Li~\etal~\cite{LiGWLXSS19} learned co-attention between the question features and appearance features, and Li~\etal~\cite{LiSGLH0G19} proposed to enhance similar co-attention with self-attention \cite{VaswaniSPUJGKP17}. 
Jin~\etal~\cite{JinZG00Z19} proposed a new fine-grained temporal attention that is capable of learning object-question interactions among frames. 
Huang~\etal~\cite{HuangCZDTG20} further extended the object-question interactions with location-aware co-attention between the object features and questions.
After Jang~\etal~\cite{JangSYKK17} first proposed the two-stream structure to use motion and appearance features simultaneously, more research works focused on learning spatio-temporal co-attention by leveraging appearance, motion, and question interactions.
To integrate the motion features, appearance features, and questions together, myriads of reasoning modules have been proposed and achieved great progress, such as heterogeneous fusion memory \cite{FanZZW0H19}, co-memory attention \cite{GaoGCN18}, hierarchical attention \cite{ZhaoYCHZ17,LeLV020}, multi-head attention \cite{KimCKZ18}, multi-step progressive attention \cite{TapaswiZSTUF16,KimMKKY19}, and graph neural networks \cite{HuangCZDTG20, ParkLS21, JiangH20}.
In our work, we comprehensively analyze some relevant methods on Causal-VidQA, providing some heuristic observations.

\section{Causal-VidQA Dataset}
\label{sec:dataset}

In this section, we will elaborate our proposed Causal-VidQA dataset, which studies both evidence reasoning and commonsense reasoning in real-world actions. 
In Sec. \ref{sec:dataset:task}, we provide a comprehensive definition for different types of questions involved in our Causal-VidQA dataset . 
In Sec. \ref{sec:dataset:construction}, we introduce the process of constructing our dataset. 
In Sec. \ref{sec:dataset:statistics}, we reveal some statistic information about our Causal-VidQA dataset.
In Sec. \ref{sec:dataset:comparison}, we compare our Causal-VidQA dataset with other VideoQA datasets.

\subsection{Task Definition}
\label{sec:dataset:task}

In our Causal-VidQA dataset, we design four different types of questions to study three video understanding tasks, including scene description (Description), evidence reasoning (Explanation) and commonsense reasoning (Prediction and Counterfactual). 
In the following sections, we will introduce them one by one.

\noindent \textbf{Description.}
The descriptive questions focus on the scene description of video clips (\eg, the places, objects, actions, order, and so on). 
The scene description is the basic task for video understanding and provides a comparison with other reasoning tasks.
The requirements for these questions are that the answers can be visibly inferred from current video clips and are objective. 
Specifically, the questions include location (\textit{where}), counting (\textit{how many}), binary choices (\textit{yes/no}), temporal relation (\textit{what ... doing when ...?}, \textit{what ... do after ...?}), and other free-form questions.
Examples can be found in Figure \ref{fig:example} (Description).

\noindent \textbf{Explanation.}
The explanatory questions aim to explain the intentions of actions and the procedures to certain targets. 
Besides, for the answers to explain the questions based on video clips, we ensure that the questions are visible events and the answers are visible clues responsible for the questions. 
Accordingly, explanatory questions can be categorized as: 1. why the objects act in a certain manner (\textit{why}); 2. how the objects achieve an observed effect  (\textit{how}), where both causes and effects are visible in the videos. 
Examples can be found in Figure \ref{fig:example} (Explanation).

\noindent \textbf{Prediction}
The predictive questions focus on predicting the future action beyond current video clips.
Future actions, as the extension of current actions, are usually determined by temporal relation between actions and motion information about objects (order of actions, movement tendency). 
Therefore, the purpose of predictive questions is to assess the model's ability of commonsense reasoning with temporal and motion information. 
Since not all clues are visible, we also require the model to give a rationale that explains why the answer is true according to commonsense knowledge.
Examples can be found in Figure \ref{fig:example} (Prediction).

\noindent \textbf{Counterfactual}
The counterfactual questions target at imagining what would happen under different conditions.
In this paper, the variance of condition is limited to concrete and real (\textit{a rope break}, \textit{his foot get slipped}) conditions instead of abstract or imaginary (\textit{his mind get lost}, \textit{a person has wing}) conditions in video clips, which emphasizes more about scene analysis and how real world works.
Due to the similar reasons in Sec. \ref{sec:dataset:task} \textbf{Prediction}, we also require the model to give a rationale that explains why its answer is true according to commonsense knowledge.
Examples can be found in Figure \ref{fig:example} (Counterfactual).

\subsection{Dataset Construction}
\label{sec:dataset:construction}

\subsubsection{Video Source}
In this work, we target at real-world video understanding, without any restriction to scenes or actions.
According to this goal, we review multiple different video datasets and find that the human action video dataset, Kinetics-700 (2020 version) \cite{KayCSZHVVGBNSZ17} matches our requirements well. 
Specifically, Kinetics-700, as the largest human action video dataset, contains 647,907 videos from 700 different human actions including single human actions, human-object interactions, and human-human interactions.
Considering that some videos are short or broken, we select 546,882 unbroken videos longer than 9s for further annotation.

\subsubsection{Annotation}
\label{sec:dataset:construction:anno}
Our annotation can be divided into three stages: instance segmentation annotation, rational video selection, and question-answer annotation. 
Specifically, instance segmentation annotation aims to assign the semantic labels to different instances in the whole video clips to facilitate the text description.
Rational video selection aims to manually select the rational videos that not only have correct segmentation labels, but also are proper to raise high quality questions.
Question-answer annotation aims to design a thorough procedure, which is capable of ensuring to propose high quality questions along with proper answers and reasons.
After these three annotation stages, we have 26,900 video clips from 666 action categories along with 107,600 questions, 107,600 answers, and 53,800 reasons.
Details about the annotation can be found in Supplementary.

\begin{figure}
\begin{center}
\includegraphics[width=1.0\linewidth]{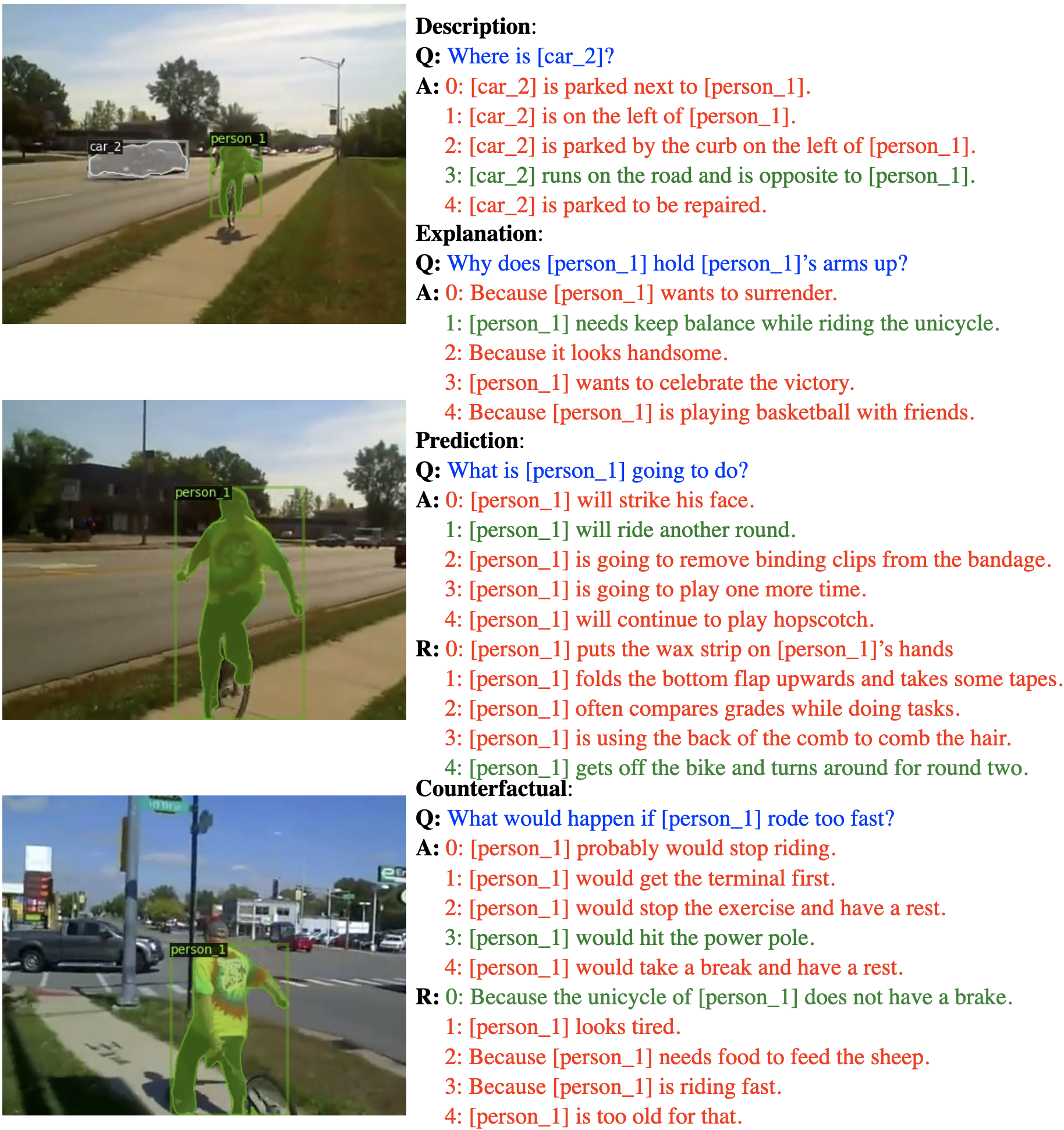}
\end{center}
   \caption{Example of multi-choice question-answering.
   The answers and reasons highlighted in bold are the correct answers and reasons.
   Best viewed by zooming in.}
\label{fig:example2}
\end{figure}

\subsubsection{Dataset Split} 
\label{sec:dataset:construction:split}
After we get the rational video clips with all the questions, answers, and reasons, the next step is to split them into training set, valid set, and test set.
Considering that the video scenes in each action category are similar, if we randomly split the videos into these three sets, the reasoning may take a short-cut by simply correlating the action categories with the questions and answers.
To prevent this situation, we randomly split the dataset into training/validation/testing set according to the action categories with a ratio of 7:1:2, resulting in 18,776 training videos, 2,695 validation videos, and 5,429 testing videos.

\subsubsection{Multi-choice Generation}
In common currency, the distractors in multi-choice QA should be different from each other, semantically coherent in answering the questions, and different from the correct answer in terms of the meaning.
Therefore, we follow the similar process in NExT-QA \cite{XiaoSYC21} to generate the multi-choice answers and reasons.
For the answers generation, the steps are:
\textbf{a)} group the questions by the question types; 
\textbf{b)} retrieve the top 50 questions similar to each question in the same question group based on the cosine similarities of sentence-level BERT \cite{DevlinCLT19} feature and regard the corresponding answer as distractor candidates;
\textbf{c)} filter the redundant and similar answers to the correct answer by 1) the lemmatized variants are the same; 2) the cosine similarity of their feature vectors is large than 0.9;
\textbf{d)} sample four qualified candidates as distracting answers;
\textbf{e)} randomly and evenly insert the correct answers to form 5 options;
\textbf{f)} manually check all the question-answer tuples to ensure that each question corresponds to one correct answer.
For the reason generation, we take similar steps as answer generation. 
The only difference is the step \textbf{b)}, in which the distractor candidates for reason are acquired by measuring the similarities of both questions and answers.
Note that we match the answers and reasons individually for training, validation and testing set to ensure that there is no question/answer overlap.
Examples are shown in Figure \ref{fig:example2} and Supplementary.

\begin{figure}
\begin{center}
\includegraphics[width=1.0\linewidth]{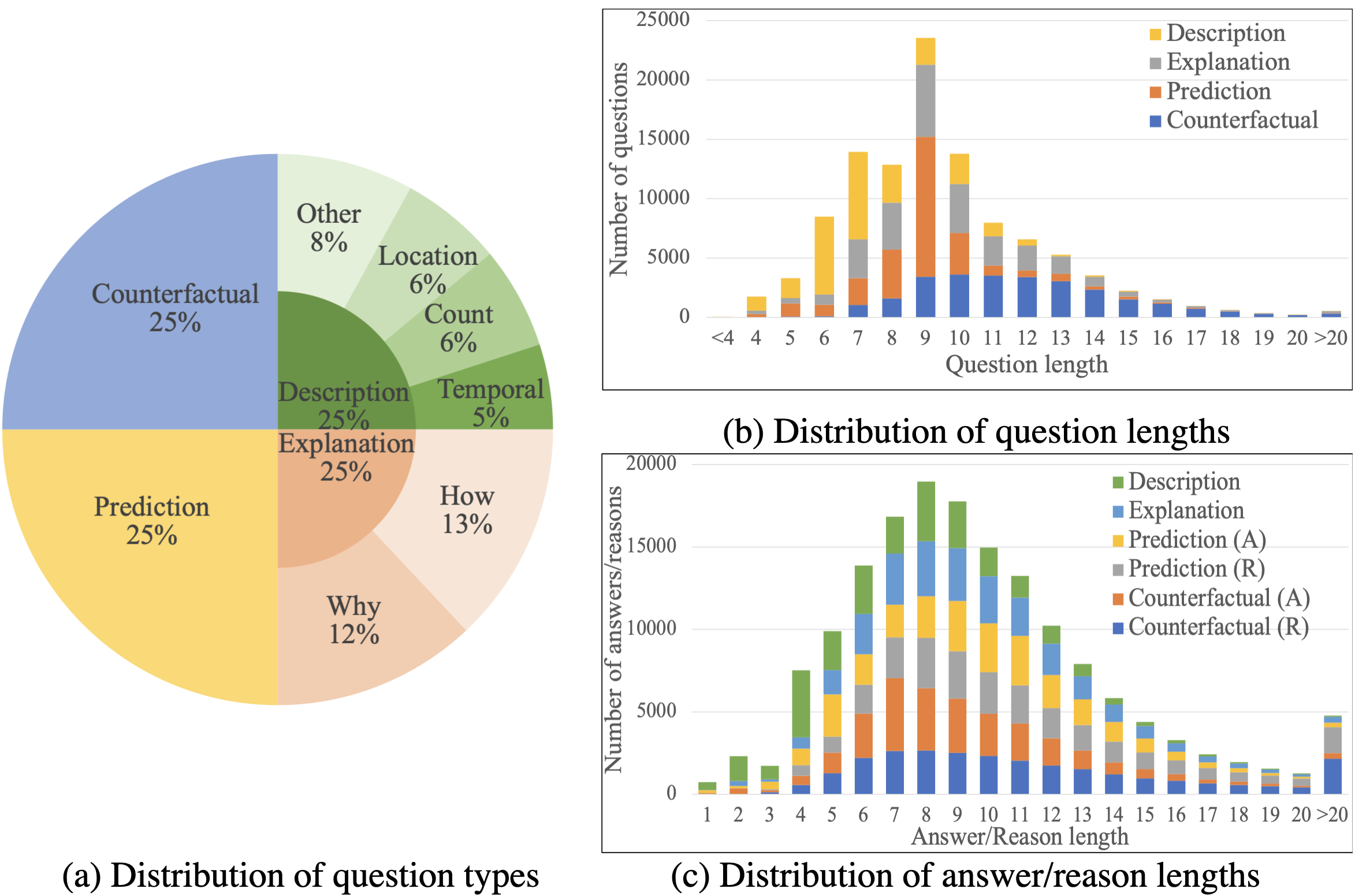}
\end{center}
   \caption{Data statistics. 
   (a) Distribution of the question types. 
   (b) The average question length is 9.5, and the specific lengths for description, explanation, prediction and counterfactual questions are 7.5, 9.6, 8.9 and 11.9 respectively. 
   (c) The average answer/reason length is 9.8. Specific lengths for description, explanation, prediction(A), prediction(R), counterfactual(A) and counterfactual(R) are 7.3, 9.6, 9.6, 11.3, 9.2 and 11.5 respectively.}
\label{fig:statistics}
\end{figure}

\subsection{Dataset Statistics}
\label{sec:dataset:statistics}

Causal-VidQA contains 26,900 video clips, including 18,776 for training, 2,695 for validation and 5,429 for testing. 
For each video clip in our Causal-VidQA dataset, it has four different types of questions, including description, explanation, prediction, and counterfactual. 
The distribution of the questions and answers are shown in Figure \ref{fig:statistics}. 
From Figure \ref{fig:statistics} (a), we can find that the temporal, location, count, and free-form questions in description have similar ratios.
Besides, in predictive questions, the questions starting with `why' and `how' constitute 13\% and 12\% respectively.

The distribution of question word and answer word length is shown in Figure \ref{fig:statistics} (b).
Generally speaking, the average length of our questions is 9.5 words and is shorter than that in NExT-QA \cite{XiaoSYC21} (11.6), which is mainly because that we use the segmentation label to indicate each instance in video clips.
Besides, the descriptive questions are the shortest while the counterfactual questions are the longest.
Because the counterfactual questions usually need more words to describe extra conditions.
Comparing the answers and reasons, the answers for descriptive questions are also the shortest while both types of reasons are the longest, which is due to that the reasons contain relative complex commonsense to clarify why the answer is true.

\subsection{Dataset Comparison}
\label{sec:dataset:comparison}

Compared with other datasets, our Causal-VidQA has several unique properties (detailed comparison is in Supplementary.). 
First, Causal-VidQA is the first dataset that combines scene description, evidence reasoning and commonsense reasoning in VideoQA.
Second, for each video clips, we assign pixel-level segmentation and object label to the main instances, which not only makes the task focus on the reasoning but also enriches the visual-text interaction. 
Finally, the video clips in Causal-VidQA are all from the Kinetics-700 including 666 action categories, which are rich and diverse in terms of objects and actions, and are closely related to daily life. 
This is different from the popular TVQA \cite{LeiYBB18}, MovieQA \cite{TapaswiZSTUF16}, and CLEVRER \cite{YiGLK0TT20} dataset that are based on some unrealistic scenes (\eg, the TV shows, movie clips, game engine).

\section{Experiments}
\label{sec:experiment}


\subsection{Experiment Setting}
\label{sec:experiment:setting}

\noindent \textbf{Evaluation.}
Considering that the answers and reasons are both formed as multi-choice QA, we report the accuracy for all four types of questions.
Note that, for the predictive and counterfactual questions, each question has the candidate answers and reasons. 
Therefore, ``correctness" is defined as choosing the right answer and reason at the same time.

\noindent \textbf{Configuration.}
On the video side, we uniformly sample 8 segments for each video clip, and each segment has 16 consecutive frames.
For the frame-level representation, we employ ResNet-101 \cite{HeZRS16} pre-trained on ImageNet \cite{DengDSLL009} to extract the appearance features.
For the segment-level representation, we employ inflated 3D ResNeXt-101 \cite{HaraKS18, XieGDTH17} pre-trained on Kinetics-400 \cite{KayCSZHVVGBNSZ17} to extract the motion features.
Huang~\etal~\cite{HuangCZDTG20} represents the video with the frame-level appearance features and object features. 
Therefore, we also employ Faster-RCNN \cite{RenHGS15} pre-trained on Visual Genome \cite{KrishnaZGJHKCKL17} to extract the object features.
On the language side, we study both GloVe \cite{PenningtonSM14} and off-the-shelf BERT \cite{DevlinCLT19} for words/tokens representation.
Besides, for segmentation labels in text, like ``[person\_1]'' in Figure \ref{fig:example}, we first employ the same Faster-RCNN to extract the object features of the corresponding instance among all frames, and then average them as the segmentation label feature vectors.
Next, the segmentation label feature vectors are mapped into the same dimension as text embedding and then added to the corresponding word/token embeddings, where the parameters for mapping are learned in training.
During training and inference, the candidate answers and reasons are concatenated to the questions, and hinge loss is applied to maximize the margins between correct and incorrect pairs.

\begin{table}[]
\setlength{\tabcolsep}{1.3mm}{%
\resizebox{0.47\textwidth}{!}{%
\begin{tabular}{ccccccc}
\hline
Settings & Text Feature & $Acc_D$ & $Acc_E$ & $Acc_P$ & $Acc_C$ & $Acc$ \\ \hline
Random   & -            &  20.07  &  20.08  & 3.99    &  4.02   & 12.04 \\
Longest  & -            &  18.92  &  20.26  & 4.30    &  3.90   & 11.84 \\
Shortest & -            &  19.22  &  19.52  & 4.38    &  4.79   & 11.98 \\
Sim-AA    & BERT        &  19.11  &  18.92  & 3.97    &  4.01   & 11.50 \\
Dissim-AA & BERT        &  21.41  &  19.81  & 3.97    &  3.78   & 12.24 \\
Sim-QA    & BERT        &  20.15  &  18.55  & 4.42    &  4.39   & 11.88 \\
Dissim-QA & BERT        &  20.41  &  19.81  & 4.16    &  3.74   & 12.03 \\ \hline
\end{tabular}%
}
}
\caption{Results of diagnosis settings on test set of Causal-VidQA. 
$Acc_D$, $Acc_E$, $Acc_P$, and $Acc_C$ denote accuracy for description, explanation, prediction and counterfactual question respectively.}
\label{tab:diagnosis}
\end{table}

\begin{table*}[]
\setlength{\tabcolsep}{3mm}{%
\resizebox{\textwidth}{!}{%
\begin{tabular}{cccccccccccc}
\hline
\multicolumn{1}{c}{\multirow{2}{*}{Method}} & \multicolumn{1}{c}{\multirow{2}{*}{Text Feature}} & \multirow{2}{*}{Video Feature} & \multicolumn{1}{c}{\multirow{2}{*}{$Acc_D$}} & \multicolumn{1}{c}{\multirow{2}{*}{$Acc_E$}} & \multicolumn{3}{c}{$Acc_P$} & \multicolumn{3}{c}{$Acc_C$} & \multicolumn{1}{c}{\multirow{2}{*}{$Acc$}} \\ \cline{6-11}
\multicolumn{1}{c}{} & \multicolumn{1}{c}{} &  & \multicolumn{1}{c}{} & \multicolumn{1}{c}{} & Q $\rightarrow$ A & Q $\rightarrow$ R & Q $\rightarrow$ AR & Q $\rightarrow$ A & Q $\rightarrow$ R & Q $\rightarrow$ AR & \multicolumn{1}{c}{} \\ \hline
BlindQA \cite{HochreiterS97} & GloVe & - & 38.66 & 30.58 & 28.68 & 30.48 & 13.91 & 21.29 & 21.61 & 6.56 & 22.43 \\
EVQA \cite{AntolALMBZP15} & GloVe & App.+Mot. & 42.88 & 38.29 & 36.89 & 36.23 & 18.29 & 27.72 & 27.57 & 10.63 & 27.52 \\
CoMem \cite{GaoGCN18} & GloVe & App.+Mot. & 59.26 & 54.23 & 43.93 & 45.37 & 26.32 & 42.97 & 42.24 & 22.25 & 40.51 \\
HME \cite{FanZZW0H19} & GloVe & App.+Mot. & 47.25 & 43.80 & 41.02 & 42.53 & 23.25 & 35.29 & 34.19 & 15.34 & 32.41 \\
HCRN \cite{LeLV020} & GloVe & App.+Mot. & 58.89 & 53.53 & 43.14 & 45.07 & 26.17 & 43.69 & 43.47 & 22.75 & 40.33 \\
HGA \cite{JiangH20} & GloVe & App.+Mot. & 60.32 & 55.02 & 46.55 & 47.21 & 28.53 & 44.00 & 44.04 & 23.63 & 41.88 \\
B2A \cite{ParkLS21} & GloVe & App.+Mot. & 61.29 & 56.43 & 46.82 & 48.17 & 30.01 & 45.12 & 44.99 & 25.29 & 43.26 \\ \hline
BlindQA \cite{HochreiterS97} & BERT & - & 60.78 & 59.46 & 44.01 & 45.73 & 26.81 & 47.97 & 49.54 & 28.71 & 43.94 \\
EVQA \cite{AntolALMBZP15} & BERT & App.+Mot. & 63.73 & 60.95 & 45.68 & 46.40 & 27.19 & 48.96 & 51.46 & 30.19 & 45.51 \\
CoMem \cite{GaoGCN18} & BERT & App.+Mot. & 64.08 & 62.79 & 51.00 & 50.36 & 31.41 & 51.61 & 53.10 & 32.55 & 47.71 \\
HME \cite{FanZZW0H19} & BERT & App.+Mot.& 63.36 & 61.45 & 50.29 & 47.56 & 28.92 & 50.38 & 51.65 & 30.93 & 46.16  \\
HCRN \cite{LeLV020} & BERT & App.+Mot. & 65.35 & 61.61 & \textbf{51.74} & \textbf{51.26} & \textbf{32.57} & 51.57 & 53.44 & 32.66 & 48.05 \\
HGA \cite{JiangH20} & BERT & App.+Mot. & 65.67 & \textbf{63.51} & 49.36 & 50.62 & 32.22 & 52.44 & 55.85 & 34.28 & 48.92 \\
B2A \cite{ParkLS21} & BERT & App.+Mot. & \textbf{66.21} & 62.92 & 48.96 & 50.22 & 31.15 & \textbf{53.27} & \textbf{56.27} & \textbf{35.16} & \textbf{49.11} \\ \hline
Human & - & - & 95.24 & 94.74 & 92.38 & 93.66 & 89.30 & 93.89 & 92.77 & 90.05 & 92.33 \\ \hline
\end{tabular}%
}
}
\caption{Results of existing methods and human on test set of Causal-VidQA. 
App., Mot., and Obj. represent the appearance feature, motion feature, and object feature mentioned in Sec. \ref{sec:experiment:setting}.
The best result from model is highlighted in bold.}
\label{tab:main results}
\end{table*}

\subsection{Dataset Diagnosis}
\label{sec:experiment:diagnosis}

In this section, we design some special setting in Table~\ref{tab:diagnosis}, including random selection, length selection, and similarity-based selection to diagnose some potential biases that might appear in our Causal-VidQA dataset.

\noindent \textbf{Random Selection.}
In this setting, we randomly select one candidate as the chosen answer/reason and then calculate the accuracy among all four types of questions.
From the \textit{Random} line in Table~\ref{tab:diagnosis}, the accuracy of random selection is 20\% (\emph{resp.}, 4\%) for descriptive and explanatory (\emph{resp.}, for predictive and counterfactual) questions, which indicates that the correct answers and reasons are uniformly distributed among all options. 

\noindent \textbf{Length Selection.}
In this setting, we target at evaluating whether the correct answer has length bias, that is, the longest or shortest candidate has higher probability as the correct answer/reason. 
Therefore, we evaluate two settings in this part, Longest and Shortest, which means always choosing the longest or shortest answer/reason.
From the \textit{Longest} and \textit{Shortest} lines in Table~\ref{tab:diagnosis}, the accuracy for descriptive and explanatory (\emph{resp.}, for predictive and counterfactual) questions is close to 20\% (\emph{resp.}, 4\%), therefore, the models cannot achieve satisfied results on Causal-VidQA directly by the length of answers/reasons.

\noindent \textbf{Similarity-based Selection.}
In this setting, we aim to evaluate whether the sentence-level BERT \cite{DevlinCLT19} feature used in multi-choose generation introduces feature bias, for example, the correct answer/reason has the closest or furthest distance with questions or other options.
Therefore, we validate four feature-based retrieval settings (\ie, Sim-AA, Dissim-AA, Sim-QA, Dissim-QA), where Sim-AA (\emph{resp.} Dissim-AA) denotes that the answer/reason has the closest (\emph{resp.} furthest) cosine distance with other options are selected as the chosen one, and Sim-QA (\emph{resp.} Dissim-QA) denotes that the answer/reason has the closest (\emph{resp.} furthest) cosine distance with the question are selected as the chosen one.
From the \textit{Sim-AA}, \textit{Dissim-AA}, \textit{Sim-QA}, and \textit{Dissim-QA} lines in Table~\ref{tab:diagnosis}, the accuracy for descriptive and explanatory (\emph{resp.}, for predictive and counterfactual) questions is close to 20\% (\emph{resp.}, 4\%), which indicates that the correct answer and proper reason cannot be deduced simply based on semantic similarity among answers or between the questions and answers.

\subsection{Existing Methods}
\label{sec:experiment:main}

In this section, we evaluate our Causal-VidQA dataset with several existing VideoQA in Table \ref{tab:main results}, which include BlindQA \cite{HochreiterS97}, EVQA \cite{AntolALMBZP15}, CoMem \cite{GaoGCN18}, HME \cite{FanZZW0H19}, HGA \cite{JiangH20}, HCRN \cite{LeLV020}, and B2A \cite{ParkLS21}.

Before evaluating the models specifically designed for VideoQA, we first study a blind version baseline \cite{HochreiterS97} by considering the question-answers only and ignoring the video clips. 
For this baseline, we model the concatenation of question and each answer/reason with LSTM to get the text vector, during which the words/tokens representations are initialized with either GloVe \cite{PenningtonSM14} or off-the-shelf BERT \cite{DevlinCLT19}.
Then we also evaluate the EVQA \cite{AntolALMBZP15}, which is a simple extension of BlindQA by processing video feature with another LSTM and then adding the video vector to the text vector for answer prediction.
Comparing the performance between BlindQA \cite{HochreiterS97} and EVQA \cite{AntolALMBZP15} with different text features, we can find that EVQA \cite{AntolALMBZP15} improves BlindQA baseline quite significant, since the visual information is quite important to guide the question answering.

To further enhance the visual-language interaction, CoMem \cite{GaoGCN18} applies additional spatio-temporal attention modules with memory bank to fuse the appearance and motion features with the text features. 
Based on the model architecture of CoMem \cite{GaoGCN18}, HME \cite{FanZZW0H19} extend the spatio-temporal attention with memory module by a multi-cycle interaction among appearance, motion and text features. 
From Table \ref{tab:main results}, we can also find that CoMem \cite{JangSYKK17} and HME \cite{FanZZW0H19} improve the performance of EVQA \cite{AntolALMBZP15} by a large margin in all four types of questions, especially in description.

HCRN \cite{LeLV020} is a hierarchical model with the proposed conditional relation networks (CRN) to aggregate visual and language information in two stages. 
In the first stage, CRN blocks aim to aggregate the frame-level appearance features conditioned on the segment-level motion features and text features as the segment representation.
In the second stage, CRN blocks further aggregate segment representation conditioned on the video-level motion features and text features.
The stage-wise feature aggregation captures multi-granularity text and motion features in a coarse-to-fine manner.
In Table~\ref{tab:main results}, we can also find that HCRN achieves comparable results on all four types of questions, especially in predictive reasoning.

Graph Convolution Network (GCN) is employed among HGA \cite{JiangH20} and B2A \cite{ParkLS21} for inter- and intra-modality interaction and visual-language reasoning. 
Specifically, HGA \cite{JiangH20} first utilizes co-attention to align visual representation (motion and appearance) and the language representation, and then applies GCN by regarding the feature in both visual and language modality into a same heterogeneous graph.
B2A \cite{ParkLS21} combines the inter- and intra-modality interaction together by first build graph within text features, motion features, and appearance features. 
Then, B2A \cite{ParkLS21} employs question-to-visual interaction and visual-to-visual interaction to get the final text, motion, and appearance representation.
As shown in Table \ref{tab:main results}, HGA \cite{JiangH20} and B2A \cite{ParkLS21} show superior performance among all four types of questions under both types of text features \footnote{More experiments about ablation study are in Supplementary.}.

Furthermore, we can also find that BERT remarkably boosts the results over GloVe, which indicates that BERT representation can provide richer semantic information in language side and prevent overfitting. Besides, comparing the predictive and counterfactual questions, we can find that BERT improves more on the counterfactual question, which is because the ``Next Sentence Prediction'' (NSP) task used in BERT pre-training can integrate commonsense knowledge and is helpful for commonsense reasoning.

\begin{figure}
\begin{center}
\includegraphics[width=1.0\linewidth]{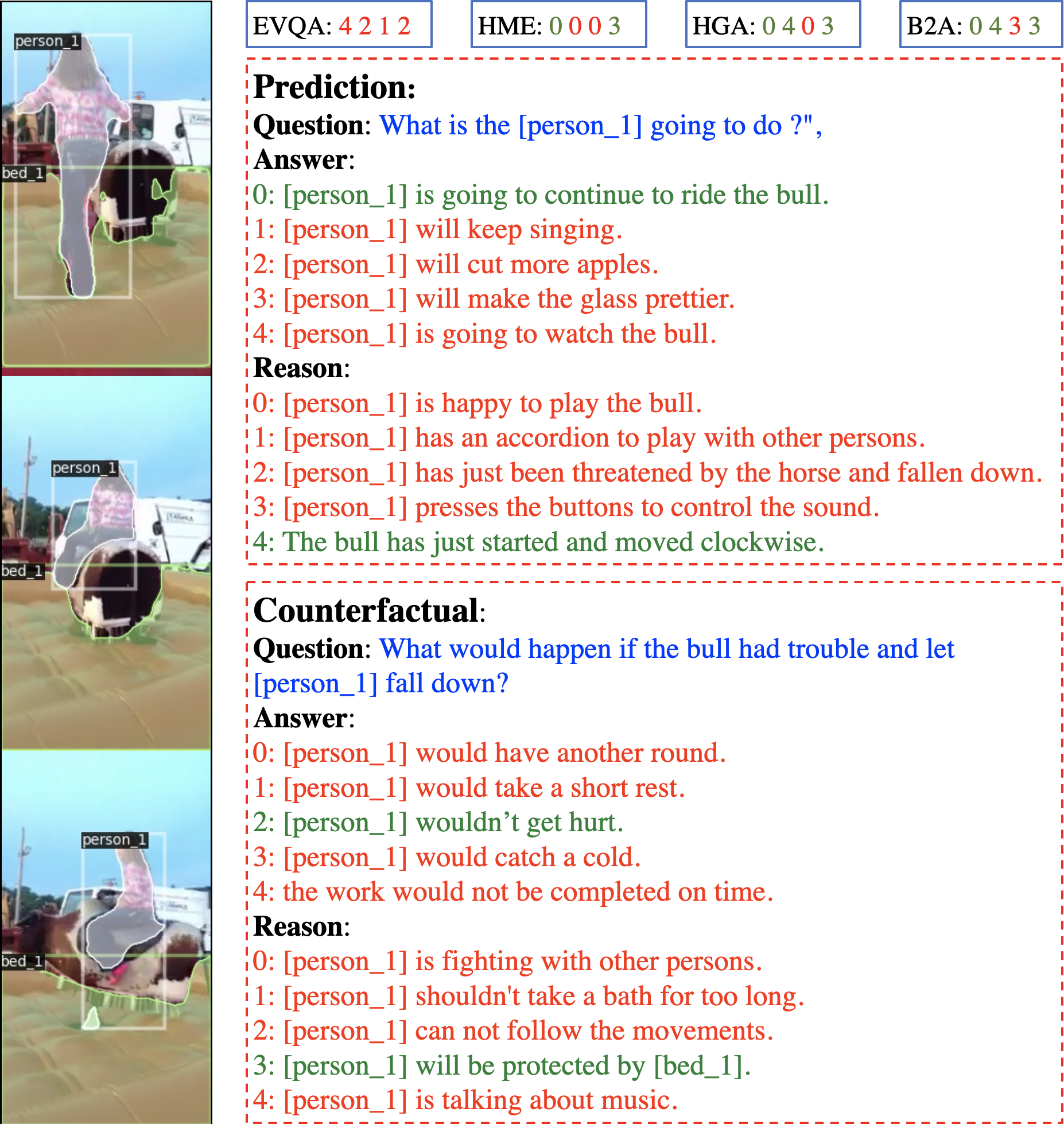}
\end{center}
   \caption{Qualitative examples of EVQA \cite{AntolALMBZP15}, HME \cite{FanZZW0H19}, HGA \cite{JiangH20}, B2A \cite{ParkLS21} on predictive and counterfactual questions. 
   Numbers in blue boxes indicates the choice of model for the prediction-answer, prediction-reason, counterfactual-answer, and counterfactual-reason, where the red (\emph{resp.}, green) number represents that the choice of the method is wrong (\emph{resp.}, right). 
   Correct answers and reasons are highlighted in bold.}
\label{fig:modelcase}
\end{figure}


\subsection{Qualitative Results}
\label{sec:experiment:qualitative}
We present qualitative examples on predictive and counterfactual questions in Figure \ref{fig:modelcase}. 
From this example, we can find that none of the model can correctly answer the counterfactual question, because it is hard for existing methods to correlate ``fall down'' with ``hurt''.
Besides, the correct reason for the counterfactual question may be achieved by simply correlating the ``[bed\_1]'' in both video and language.
We further replace the segmentation label ``[bed\_1]'' with word ``bed'', and find that HME, HGA, and B2A all give wrong reason to this question.
This reveals a limitation of multi-choice question-answering in our dataset, \ie, the answer or reason might be chosen by correlating segmentation label in video and text.
However, considering the improvement of HGA over BlindQA in all types of questions, this problem would also not dominate the performance.

\section{Discussion and Limitation}

In this section, we discuss about current solutions, future directions and limitations of our dataset.

\noindent \textbf{Discussion.} For the language representation, we can find that the off-the-shelf BERT \cite{DevlinCLT19} representation outperforms GloVe \cite{PenningtonSM14} embedding by a large margin, especially in evidence reasoning and commonsense reasoning.
On the one hand, this phenomenon shows that the pre-trained BERT has great ability for language representation. 
On the other hand, it also indicates that some Causal relation within the language have been modeled by the language pre-training \cite{DevlinCLT19}.
Therefore, in the future, we highly suggest that the pre-trained language models, like ELMo \cite{PetersNIGCLZ18}, BERT \cite{DevlinCLT19}, ALBERT \cite{LanCGGSS20}, \emph{etc.}, should be further explored in this field.

For the video representation, existing works have proven the value of combining motion and appearance feature.
However, the use of both appearance and motion features are still limited to attention or graph interaction, which is not enough to fully exploit the information within these two kind of features.
Besides, the object features have shown their ability in ImageQA \cite{ZellersBFC19}, whereas, it has not attracted enough attention in VideoQA, because the video scenes in current datasets are relatively simple and the object-level motion feature is also hard to extract.
In the future, extending the appearance and motion feature to object-level appearance and motion feature might be essential for video representation in VideoQA.

For the cross-modality interaction, current solutions have explored the attention-based solutions \cite{JangSYKK17}, memory-based solutions \cite{FanZZW0H19}, and graph-based solutions \cite{JiangH20, ParkLS21}. 
Currently speaking, graph-based solutions, especially the sparse graph \cite{ParkLS21}, are superior for all types of questions, however, the performance still has a large gap with human (47.46\% V.S. 92.33\%) and thus offers ample opportunity for further improvement.
Apart from the solutions mentioned before, more new technique should be further exploited in the future.
For example, the causal structure graph could bring causal prior from human and knowledge base may introduce richer background knowledge for reasoning.

\noindent \textbf{Limitation.}
As we have discussed and analyzed in Sec.\ref{sec:experiment:qualitative}, the format of multi-choice question answering suffers from that the answer or reason might be chosen by correlating segmentation labels between videos and texts, however, this phenomenon is not serious in our Causal-VidQA dataset.

Besides, for the predictive and the counterfactual questions, the answers/reasons for these questions are subjective. 
Although we ensure that at least 5 different persons achieve agreement for all distractors, the distractors may still not satisfy the logic of everyone.

Furthermore, our dataset only supports multi-choice QA, because for predictive and counterfactual questions, there might be more than one correct answers.
In the future, extending predictive and counterfactual questions to reason-based open-ended questions may be a feasible solution.

\section{Conclusion}
\label{sec:conclusion}
We have explored evidence and commonsense reasoning in video question-answering to advance video understanding towards deeper reasoning.
Therefore, we have contributed Causal-VidQA, a new challenging video question-answering benchmark containing four types of questions.
We have extensively evaluated some SOTA video question-answering methods on our Causal-VidQA, providing detailed comparison and in-depth analyses for future research.

\section*{Acknowledgement} The work was supported by the National Key R\&D Program of China (2018AAA0100704), the National Science Foundation of China (62076162), and the Shanghai Municipal Science and Technology Major Project, China (2021SHZDZX0102, 20511100300). We thank Wu Wen Jun Honorary Doctoral Scholarship, AI Institute, Shanghai Jiao Tong University. We thank all annotators for their remarkable work in data annotation.

{\small
\bibliographystyle{ieee_fullname}
\bibliography{main}

\begin{thebibliography}{10}\itemsep=-1pt

\bibitem{AntolALMBZP15}
Stanislaw Antol, Aishwarya Agrawal, Jiasen Lu, Margaret Mitchell, Dhruv Batra,
  C.~Lawrence Zitnick, and Devi Parikh.
\newblock {VQA:} visual question answering.
\newblock In {\em ICCV 2015}, pages 2425--2433, 2015.

\bibitem{buchsbaum2015inferring}
Daphna Buchsbaum, Thomas~L Griffiths, Dillon Plunkett, Alison Gopnik, and Dare
  Baldwin.
\newblock Inferring action structure and causal relationships in continuous
  sequences of human action.
\newblock {\em Cognitive psychology}, 76:30--77, 2015.

\bibitem{CarreiraZ17}
Jo{\~{a}}o Carreira and Andrew Zisserman.
\newblock Quo vadis, action recognition? {A} new model and the kinetics
  dataset.
\newblock In {\em CVPR 2017}, pages 4724--4733, 2017.

\bibitem{DengDSLL009}
Jia Deng, Wei Dong, Richard Socher, Li{-}Jia Li, Kai Li, and Li Fei{-}Fei.
\newblock Imagenet: {A} large-scale hierarchical image database.
\newblock In {\em CVPR 2009}, pages 248--255, 2009.

\bibitem{DevlinCLT19}
Jacob Devlin, Ming{-}Wei Chang, Kenton Lee, and Kristina Toutanova.
\newblock {BERT:} pre-training of deep bidirectional transformers for language
  understanding.
\newblock In {\em NAACL-HLT 2019}, pages 4171--4186, 2019.

\bibitem{FanZZW0H19}
Chenyou Fan, Xiaofan Zhang, Shu Zhang, Wensheng Wang, Chi Zhang, and Heng
  Huang.
\newblock Heterogeneous memory enhanced multimodal attention model for video
  question answering.
\newblock In {\em CVPR 2019}, pages 1999--2007, 2019.

\bibitem{FangGBBY20}
Zhiyuan Fang, Tejas Gokhale, Pratyay Banerjee, Chitta Baral, and Yezhou Yang.
\newblock Video2commonsense: Generating commonsense descriptions to enrich
  video captioning.
\newblock In {\em EMNLP 2020}, pages 840--860, 2020.

\bibitem{Feichtenhofer0M19}
Christoph Feichtenhofer, Haoqi Fan, Jitendra Malik, and Kaiming He.
\newblock Slowfast networks for video recognition.
\newblock In {\em ICCV 2019}, pages 6201--6210, 2019.

\bibitem{GaoGCN18}
Jiyang Gao, Runzhou Ge, Kan Chen, and Ram Nevatia.
\newblock Motion-appearance co-memory networks for video question answering.
\newblock In {\em CVPR 2018}, pages 6576--6585, 2018.

\bibitem{GuadarramaKMVMDS13}
Sergio Guadarrama, Niveda Krishnamoorthy, Girish Malkarnenkar, Subhashini
  Venugopalan, Raymond~J. Mooney, Trevor Darrell, and Kate Saenko.
\newblock Youtube2text: Recognizing and describing arbitrary activities using
  semantic hierarchies and zero-shot recognition.
\newblock In {\em ICCV 2013}, pages 2712--2719, 2013.

\bibitem{HaraKS18}
Kensho Hara, Hirokatsu Kataoka, and Yutaka Satoh.
\newblock Can spatiotemporal 3d cnns retrace the history of 2d cnns and
  imagenet?
\newblock In {\em CVPR 2018}, pages 6546--6555, 2018.

\bibitem{HeZRS16}
Kaiming He, Xiangyu Zhang, Shaoqing Ren, and Jian Sun.
\newblock Deep residual learning for image recognition.
\newblock In {\em CVPR 2016}, pages 770--778, 2016.

\bibitem{HeilbronEGN15}
Fabian~Caba Heilbron, Victor Escorcia, Bernard Ghanem, and Juan~Carlos Niebles.
\newblock Activitynet: {A} large-scale video benchmark for human activity
  understanding.
\newblock In {\em CVPR 2015}, pages 961--970, 2015.

\bibitem{HendricksWSSDR17}
Lisa~Anne Hendricks, Oliver Wang, Eli Shechtman, Josef Sivic, Trevor Darrell,
  and Bryan~C. Russell.
\newblock Localizing moments in video with natural language.
\newblock In {\em ICCV 2017}, pages 5804--5813, 2017.

\bibitem{HochreiterS97}
Sepp Hochreiter and J{\"{u}}rgen Schmidhuber.
\newblock Long short-term memory.
\newblock {\em Neural Comput.}, 9(8):1735--1780, 1997.

\bibitem{HuangCZDTG20}
Deng Huang, Peihao Chen, Runhao Zeng, Qing Du, Mingkui Tan, and Chuang Gan.
\newblock Location-aware graph convolutional networks for video question
  answering.
\newblock In {\em AAAI 2020}, pages 11021--11028, 2020.

\bibitem{JangSYKK17}
Yunseok Jang, Yale Song, Youngjae Yu, Youngjin Kim, and Gunhee Kim.
\newblock {TGIF-QA:} toward spatio-temporal reasoning in visual question
  answering.
\newblock In {\em CVPR 2017}, pages 1359--1367, 2017.

\bibitem{JiangH20}
Pin Jiang and Yahong Han.
\newblock Reasoning with heterogeneous graph alignment for video question
  answering.
\newblock In {\em AAAI 2020}, pages 11109--11116, 2020.

\bibitem{JinZG00Z19}
Weike Jin, Zhou Zhao, Mao Gu, Jun Yu, Jun Xiao, and Yueting Zhuang.
\newblock Multi-interaction network with object relation for video question
  answering.
\newblock In {\em ACM MM 2019}, pages 1193--1201, 2019.

\bibitem{KayCSZHVVGBNSZ17}
Will Kay, Jo{\~{a}}o Carreira, Karen Simonyan, Brian Zhang, Chloe Hillier,
  Sudheendra Vijayanarasimhan, Fabio Viola, Tim Green, Trevor Back, Paul
  Natsev, Mustafa Suleyman, and Andrew Zisserman.
\newblock The kinetics human action video dataset.
\newblock {\em arXiv}, 2017.

\bibitem{KimMKKY19}
Junyeong Kim, Minuk Ma, Kyungsu Kim, Sungjin Kim, and Chang~D. Yoo.
\newblock Progressive attention memory network for movie story question
  answering.
\newblock In {\em CVPR 2019}, pages 8337--8346, 2019.

\bibitem{KimCKZ18}
Kyung{-}Min Kim, Seong{-}Ho Choi, Jin{-}Hwa Kim, and Byoung{-}Tak Zhang.
\newblock Multimodal dual attention memory for video story question answering.
\newblock In {\em ECCV 2018}, pages 698--713, 2018.

\bibitem{KrishnaHRFN17}
Ranjay Krishna, Kenji Hata, Frederic Ren, Li Fei{-}Fei, and Juan~Carlos
  Niebles.
\newblock Dense-captioning events in videos.
\newblock In {\em ICCV 2017}, pages 706--715, 2017.

\bibitem{KrishnaZGJHKCKL17}
Ranjay Krishna, Yuke Zhu, Oliver Groth, Justin Johnson, Kenji Hata, Joshua
  Kravitz, Stephanie Chen, Yannis Kalantidis, Li{-}Jia Li, David~A. Shamma,
  Michael~S. Bernstein, and Li Fei{-}Fei.
\newblock Visual genome: Connecting language and vision using crowdsourced
  dense image annotations.
\newblock {\em Int. J. Comput. Vis.}, 123(1):32--73, 2017.

\bibitem{LanCGGSS20}
Zhenzhong Lan, Mingda Chen, Sebastian Goodman, Kevin Gimpel, Piyush Sharma, and
  Radu Soricut.
\newblock {ALBERT:} {A} lite {BERT} for self-supervised learning of language
  representations.
\newblock In {\em ICLR 2020}, 2020.

\bibitem{LeLV020}
Thao~Minh Le, Vuong Le, Svetha Venkatesh, and Truyen Tran.
\newblock Hierarchical conditional relation networks for video question
  answering.
\newblock In {\em CVPR 2020}, pages 9969--9978, 2020.

\bibitem{LeiYBB18}
Jie Lei, Licheng Yu, Mohit Bansal, and Tamara~L. Berg.
\newblock {TVQA:} localized, compositional video question answering.
\newblock In {\em EMNLP 2018}, pages 1369--1379, 2018.

\bibitem{LiGWLXSS19}
Xiangpeng Li, Lianli Gao, Xuanhan Wang, Wu Liu, Xing Xu, Heng~Tao Shen, and
  Jingkuan Song.
\newblock Learnable aggregating net with diversity learning for video question
  answering.
\newblock In {\em ACM MM 2019}, pages 1166--1174, 2019.

\bibitem{LiSGLH0G19}
Xiangpeng Li, Jingkuan Song, Lianli Gao, Xianglong Liu, Wenbing Huang, Xiangnan
  He, and Chuang Gan.
\newblock Beyond rnns: Positional self-attention with co-attention for video
  question answering.
\newblock In {\em AAAI 2019}, pages 8658--8665, 2019.

\bibitem{LiSCTGJL16}
Yuncheng Li, Yale Song, Liangliang Cao, Joel~R. Tetreault, Larry Goldberg,
  Alejandro Jaimes, and Jiebo Luo.
\newblock {TGIF:} {A} new dataset and benchmark on animated {GIF} description.
\newblock In {\em CVPR 2016}, pages 4641--4650, 2016.

\bibitem{LinGH19}
Ji Lin, Chuang Gan, and Song Han.
\newblock {TSM:} temporal shift module for efficient video understanding.
\newblock In {\em ICCV 2019}, pages 7082--7092, 2019.

\bibitem{MaharajBRCP17}
Tegan Maharaj, Nicolas Ballas, Anna Rohrbach, Aaron~C. Courville, and
  Christopher~Joseph Pal.
\newblock A dataset and exploration of models for understanding video data
  through fill-in-the-blank question-answering.
\newblock In {\em CVPR 2017}, pages 7359--7368, 2017.

\bibitem{MunSJH17}
Jonghwan Mun, Paul~Hongsuck Seo, Ilchae Jung, and Bohyung Han.
\newblock Marioqa: Answering questions by watching gameplay videos.
\newblock In {\em ICCV 2017}, pages 2886--2894, 2017.

\bibitem{ParkLS21}
Jungin Park, Jiyoung Lee, and Kwanghoon Sohn.
\newblock Bridge to answer: Structure-aware graph interaction network for video
  question answering.
\newblock In {\em CVPR 2021}, pages 15526--15535, 2021.

\bibitem{PenningtonSM14}
Jeffrey Pennington, Richard Socher, and Christopher~D. Manning.
\newblock {GloVe}: Global vectors for word representation.
\newblock In {\em EMNLP 2014}, pages 1532--1543, 2014.

\bibitem{PetersNIGCLZ18}
Matthew~E. Peters, Mark Neumann, Mohit Iyyer, Matt Gardner, Christopher Clark,
  Kenton Lee, and Luke Zettlemoyer.
\newblock Deep contextualized word representations.
\newblock In {\em NAACL 2018}, pages 2227--2237, 2018.

\bibitem{RenHGS15}
Shaoqing Ren, Kaiming He, Ross~B. Girshick, and Jian Sun.
\newblock Faster {R-CNN:} towards real-time object detection with region
  proposal networks.
\newblock In {\em NeurIPS 2015}, pages 91--99, 2015.

\bibitem{SenerSY20}
Fadime Sener, Dipika Singhania, and Angela Yao.
\newblock Temporal aggregate representations for long-range video
  understanding.
\newblock In {\em ECCV 2020}, pages 154--171, 2020.

\bibitem{spelke2000core}
Elizabeth~S Spelke.
\newblock Core knowledge.
\newblock {\em American psychologist}, 55(11):1233, 2000.

\bibitem{TapaswiZSTUF16}
Makarand Tapaswi, Yukun Zhu, Rainer Stiefelhagen, Antonio Torralba, Raquel
  Urtasun, and Sanja Fidler.
\newblock Movieqa: Understanding stories in movies through question-answering.
\newblock In {\em CVPR 2016}, pages 4631--4640, 2016.

\bibitem{TranBFTP15}
Du Tran, Lubomir~D. Bourdev, Rob Fergus, Lorenzo Torresani, and Manohar Paluri.
\newblock Learning spatiotemporal features with 3d convolutional networks.
\newblock In {\em ICCV 2015}, pages 4489--4497, 2015.

\bibitem{VaswaniSPUJGKP17}
Ashish Vaswani, Noam Shazeer, Niki Parmar, Jakob Uszkoreit, Llion Jones,
  Aidan~N. Gomez, Lukasz Kaiser, and Illia Polosukhin.
\newblock Attention is all you need.
\newblock In {\em NeurIPS 2017}, pages 5998--6008, 2017.

\bibitem{VondrickOPT16}
Carl Vondrick, Deniz Oktay, Hamed Pirsiavash, and Antonio Torralba.
\newblock Predicting motivations of actions by leveraging text.
\newblock In {\em CVPR 2016}, pages 2997--3005, 2016.

\bibitem{WinterbottomXMM20}
Thomas Winterbottom, Sarah Xiao, Alistair McLean, and Noura~Al Moubayed.
\newblock On modality bias in the {TVQA} dataset.
\newblock In {\em BMVC 2020}, 2020.

\bibitem{XiaoSYTC20}
Junbin Xiao, Xindi Shang, Xun Yang, Sheng Tang, and Tat{-}Seng Chua.
\newblock Visual relation grounding in videos.
\newblock In {\em ECCV 2020}, pages 447--464, 2020.

\bibitem{XiaoSYC21}
Junbin Xiao, Xindi Shang, Angela Yao, and Tat{-}Seng Chua.
\newblock Next-qa: Next phase of question-answering to explaining temporal
  actions.
\newblock In {\em CVPR 2021}, pages 9777--9786, 2021.

\bibitem{XieGDTH17}
Saining Xie, Ross~B. Girshick, Piotr Doll{\'{a}}r, Zhuowen Tu, and Kaiming He.
\newblock Aggregated residual transformations for deep neural networks.
\newblock In {\em CVPR 2017}, pages 5987--5995, 2017.

\bibitem{XuMYR16}
Jun Xu, Tao Mei, Ting Yao, and Yong Rui.
\newblock {MSR-VTT:} {A} large video description dataset for bridging video and
  language.
\newblock In {\em CVPR 2016}, pages 5288--5296, 2016.

\bibitem{YangGWSS18}
Guangyu~Robert Yang, Igor Ganichev, Xiao{-}Jing Wang, Jonathon Shlens, and
  David Sussillo.
\newblock A dataset and architecture for visual reasoning with a working
  memory.
\newblock In {\em ECCV 2018}, pages 729--745, 2018.

\bibitem{YangFX19}
Linjie Yang, Yuchen Fan, and Ning Xu.
\newblock Video instance segmentation.
\newblock In {\em ICCV 2019}, pages 5187--5196, 2019.

\bibitem{YangW020}
Zongxin Yang, Yunchao Wei, and Yi Yang.
\newblock Collaborative video object segmentation by foreground-background
  integration.
\newblock In {\em ECCV 2020}, volume 12350, pages 332--348, 2020.

\bibitem{YiGLK0TT20}
Kexin Yi, Chuang Gan, Yunzhu Li, Pushmeet Kohli, Jiajun Wu, Antonio Torralba,
  and Joshua~B. Tenenbaum.
\newblock {CLEVRER:} collision events for video representation and reasoning.
\newblock In {\em ICLR 2020}, 2020.

\bibitem{YuXYYZZT19}
Zhou Yu, Dejing Xu, Jun Yu, Ting Yu, Zhou Zhao, Yueting Zhuang, and Dacheng
  Tao.
\newblock Activitynet-qa: {A} dataset for understanding complex web videos via
  question answering.
\newblock In {\em AAAI 2019}, pages 9127--9134, 2019.

\bibitem{0001CLTM19}
Amir Zadeh, Michael Chan, Paul~Pu Liang, Edmund Tong, and Louis{-}Philippe
  Morency.
\newblock Social-iq: {A} question answering benchmark for artificial social
  intelligence.
\newblock In {\em CVPR 2019}, pages 8807--8817, 2019.

\bibitem{ZellersBFC19}
Rowan Zellers, Yonatan Bisk, Ali Farhadi, and Yejin Choi.
\newblock From recognition to cognition: Visual commonsense reasoning.
\newblock In {\em CVPR 2019}, pages 6720--6731, 2019.

\bibitem{ZhaoYCHZ17}
Zhou Zhao, Qifan Yang, Deng Cai, Xiaofei He, and Yueting Zhuang.
\newblock Video question answering via hierarchical spatio-temporal attention
  networks.
\newblock In {\em IJCAI 2017}, pages 3518--3524, 2017.

\bibitem{ZhouZCSX18}
Luowei Zhou, Yingbo Zhou, Jason~J. Corso, Richard Socher, and Caiming Xiong.
\newblock End-to-end dense video captioning with masked transformer.
\newblock In {\em CVPR 2018}, pages 8739--8748, 2018.

\end{thebibliography}
}

\end{document}